# Address Matching Based On Hierarchical Information

**Abstract** There is evidence that address matching plays a crucial role in many areas such as express delivery, online shopping and so on. Address has a hierarchical structure, in contrast to unstructured texts, which can contribute valuable information for address matching. Based on this idea, this paper proposes a novel method to leverage the hierarchical information in deep learning method that not only improves the ability of existing methods to handle irregular address, but also can pay closer attention to the special part of address. Experimental findings demonstrate that the proposed method improves the current approach by 3.2% points.

## 1　Introduction

Address matching plays a vital role in various fields, such as online shopping, delivery takeout and so on. An address represented by words is mapped to a physical coordinate in the real world, however an address in reality usually has multiple ways of writing. Identifying whether different descriptions of address point to a same place which is called address matching, is important but challenging.

Address matching remains a lot of persistent problems, the most difficult of which are address irregularity, including typo, element missing and redundancy, and different address writings. Table 1 illustrates part of the problems in address matching.

The term 'typo' refers to a situation that a wrong character appears in the address. A notable example of typo is "沙停村 (Shating Village)" in Table 1 and the correct writing is "沙亭村 (Shating Village)". Element missing can be loosely described as some elements in the address are missing. As a good indicator for address matching, in the second row and second column, Table 1, the address misses"戈江区 (Gejiang District)". The term 'redundancy' is generally understood to mean some character is irrelevant to address, such as address in the third row, second column in which "(不要放美禾超市)(do not put in Meihe Supermarket)" is irrelevant to address. The fourth row in Table 1 shows two different ways of writing that point to the same address. Some problems are same as common problems of text matching, such as redundancy, typo, and so on, while some problems are unique to address, such as elements missing, different writings of the same address. In this study, we attempt to solve the address matching problem by treating the address as a special kind of text.

To resolve the above problems, we assume the learning of fine-grained representation of address can solve the problem of elements missing and different writing of the same address and matching address elements in hierarchies can solve the problem of element missing and redundancy. Deep learning method in SSM can gracefully resolve the irregularity of address, and the learning and using of address hierarchical information can structurally yield more address-specific features that unstructured text does not have.

According to this idea, this paper proposes a three-stage address matching method to combine the hierarchical information with existing SSM models. In the first stage, After fine-tuning a NER model, we get element resolution in address. In the second stage, we use a pre-trained model to learn the fine-grained representation of address. In the last stage, we perform matching synthesis at different address element hierarchies and the whole address.

Table 1. Problems in Address Matching

| Difficulty | Address1 | Address2 | Label |
|---|---|---|---|
| Typo | 沙亭村 1—6 社九太路自编 7 号广州赞誉化妆品有限公司 (Guangzhou Zanyv CosmeticsCo., Ltd., No.7, Jiutai Road, Group 1-6, Shating Village) | 沙停村九太公路 7 号赞誉 (Zanyv, No. 7 Jiutai Road, Shating Village) | 2 |
| Element missing | 芜湖市云鼎国际小区 2 幢 1 单元 (Unit 1, Building 2, Yunding International Community, Wuhu City) | 弋江区云鼎国际 1 幢 (Building 1, Yunding International, Yijiang District) | 2 |
| Redundancy | 清源路绿地香榭丽公馆西区 44 号楼或者福到家超市 (不要放美禾超市)(Building 44, West District of Lvdixiangxie Mansion, Qingyuan Road or Fuzhaojia Supermarket (do not put Meihe Supermarket)) | 东营路与清源路交叉口西 150 米绿地香榭丽公馆西区 (150 meters west of the intersection of Dongying Road and Qingyuan Road, West District of Lvdixiangxie Mansion) | 1 |
| Different writing of the same address | 广东省东莞市塘厦镇石鼓朝阳工业区。高朗厂 (Shigu Chaoyang Industrial Zone, Tangxia Town, Dongguan City, Guangdong Province. Gaolang Factory) | 塘厦镇石鼓田厦大道 86 号 F 栋高朗 (Gaolang, Building F, No. 86, Shigutianxia Avenue, Tangxia Town) | 2 |

Note: This is some problems in address matching.

The results demonstrate that our method achieves the better performance than the original text matching approach by a 3.2% point improvement. Our work verifies the significance of hierarchical information and generates fresh insight into how to utilize hierarchical information in address matching.

The contributions of this paper are as follows:

- We propose an idea that introducing hierarchical information into deep learning method by learning fine-grained representation of address and matching address elements in hierarchies, which is consistent with human thought.

- Both the hierarchical information and the semantic information of the address are used for address matching, which not only focuses on the local matching but also pays attention to the overall matching.

- Results demonstrate that with the proposed method, a better-performing address matching task can also be achieved.

## 2 Related Work

In this chapter, We aim to introduce some researches about SSM and address matching.

### 2.1 Semantic Sentence Matching

The method of Semantic Sentence Matching(SSM) has been instrumental in our understanding of address matching. SSM can be divided into two categories, representation-based method, and interaction-based method.

As representative of the representation-based text matching, DSSM [1] maps two texts into vectors and measures similarity by the vectors. CDSSM[2] and LSTM-DSSM[3] ameliorate DSSM[1] in terms of capturing sequential information. SBERT [4] uses BERT to get the sentence representation vector. However, the similarity depends too much on the quality of representation, and may lose the features such as morphology and syntax.

MultiGranCNN [5] obtains the matching information at four granularity levels and calculates the matching score between two sentences to obtain a matrix used in classification. MatchPyramid [6] deemed similarity matrix as a 2D image to process. The calculation process of attention determines that the interaction between words can be calculated without matrix. Using BERT[7] for SSM, [CLS] contains sentence interaction information. However, the above methods require more labeled data and take more time.

In conclusion, these SSM methods do not take into account the hierarchical structure of address, so we then focus on address matching method.

## 2.2 Address Matching

Rule-based method and machine learning method are the mainstream research directions.

The rule-based method focuses only on the hierarchy of the address. Peng[8] uses hierarchical backtracking matching via a multi-tree composed of address element. The simplest way to obtain address element is to use an existing address element dictionary[9], which is not conducive to non-standard addresses. Other researchers use probabilistic statistical models for address resolution, such as HMM[10-11], CRF[12-13], etc. After address resolution, most researchers use a decision tree consisting of matching rules[14-15].

Much of the current machine learning methods on address matching omit the differences between text and address. Comber[16] applies Word2Vec and CRF in address matching. Santos[17] uses GRUs for modeling sequential data. SGAM[18] uses Bi-LSTM and CNN to obtain global and local information and combines attention module to obtain more semantic information. Some useful models in text matching are migrated to the task, such as ESIM[19]. Chen[20] constructs a knowledge graph and calculates the similarity between the knowledge graph entity and the address entity. With the development of pre-trained models, some researchers use the pre-trained model for address matching[21].

In general, rule-based methods take into account the hierarchical structure of addresses which is limited to address resolution[24-25]. The machine learning method can effectively alleviate the irregularity while little consideration is given to the hierarchical structure of the address.

## 3 Method

As we can see in address "天津市上海路 (Tianjin City Shanghai Road)", we usually understand the address like "天津市，上海路 (Tianjin City, Shanghai Road)" in mind. We hope that address matching can have the same ability. The first is to make the model learn the fine-grained representation of address, and the second is to compare at different hierarchies。

We propose a new method based on a pre-trained model, which can learn more hierarchical information in the address than other machine learning methods and can deal with irregularity compared with methods based on rules. This method is a pipeline structure: resolute address element, learn fine-grained representation of address, and match address in different hierarchies. The framework is as shown as Fig.1.

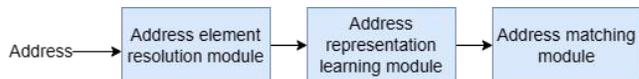

Fig.1. Cursory model.

The first module aims at get address element as hierarchical information, and the other two modules are designed to use hierarchical information.

### 3.1 Address element resolution

The address element resolution module fine-tunes a pre-train model by using the existing address element annotation data and predicts address data involved in the address matching task by the model. The specific structure of the module with example is shown in Fig.2.

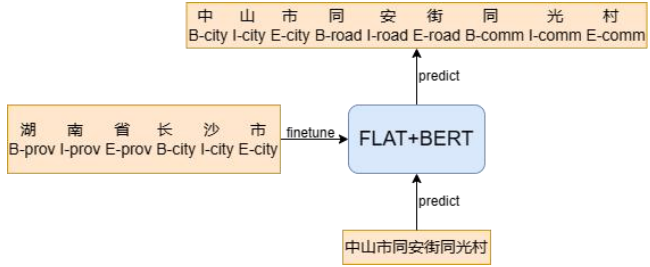

Fig.2. Structure of address element resolution module.

In order to compare the predicted address elements later, we not only need to know the segment of the address, but also the hierarchy to which the element belongs. So we decide employ and fine-tune FLAT[23], a Chinese NER Using Flat-Lattice Transformer, as the address element resolution model. To fine-tune FLAT, we must first transform address data with elements at various hierarchies into a form of a character with a label, as shown in Fig.2. In order to determine the hierarchical label and the position within the hierarchy for each character in each address, the address from the address matching dataset is then entered into the address element resolution model. In this module, we try to predict the 21 types of hierarchical label such as province, city, district, poi and so on of the address in the address matching dataset, which will be used in the following two modules.

### 3.2 Address representation learning

The address representation learning module is designed to use a pre-trained model to learn the fine-grained representation of addresses. The specific structure of the module with example is shown in Fig.3.

The composition of address is different from that of unstructured text. Each address is equivalent to a sequence combination of address elements, and there are certain relationships within and between elements, for example, the three characters in '朝阳区 (Chaoyang District)' always appear together. We choose a mask language model NEZHA[22] which is pre-trained on Chinese corpus by using the whole word mask (WWM) strategy. We hope that the model can better represent the address by focusing on learning the fine-grained address elements. We adopt the WWM mask strategy for masking the whole address element to make the model learn the representation of different hierarchies of address elements. Taking Fig.3 as an example, this module uses the output data of the previous module, processes the address into the form of segmentation according to the hierarchy, and then randomly mask different elements of address.

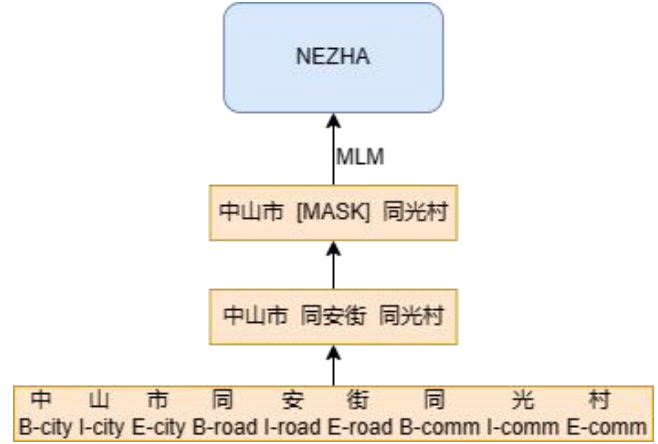

Fig.3. Structure of address representation learning module.

In this module we aim at getting a model which can learn the relationship between the address hierarchies and the internal information of the hierarchy for the next module to use.

### 3.3 Address matching

In this module, based on interactive-based text matching method, we integrate sentence-level and element-level matching for address matching. The specific structure is shown in Fig.4.

We use the address element resolution data obtained from the first module to extract different levels of address elements for splicing. The spliced address ele-ments in different levels and the address are encoded by the address expression model obtained in the previous module, and then input into different feature extraction units respectively. Finally, the vectors obtained by the feature extraction unit are concatenated and classified by the classifier.

In this module, we not only consider the matching degree between hierarchical address elements, but also consider the matching degree of the whole address text. The hierarchical elements matching takes advantage of the address-specific features, and the whole text matching takes into account the error of the address element resolution, the lack of the hierarchical elements of the address, and the text characteristics of address. While introducing hierarchical information, we avoid errors accumulation caused by address element extraction.

## 4 Experiment

### 4.1 Dataset

The data required in the experiment comes from two datasets. One is called element resolution dataset with element resolution for fine-tuning the address element resolution model in the address feature resolution module. The other is called address matching dataset that is used in the whole address matching model.

The two datasets are both provided by CCKS2021. The element resolution dataset has about 12000 train addresses and 2500 dev addresses. We continue to use the address hierarchy predefined by the address element resolution task in CCKS2021, with a total of 21 levels. The address matching dataset has about 118000 pairs of train addresses and 70000 pairs of test addresses. In the address matching dataset, label 0 means the two addresses do not match at all, label 1 means the two addresses partially match, i.e. one is a subordinate address of the other, label 2 means the two addresses match exactly, i.e. they point to the same physical location.

### 4.2 Baseline

We use the mainstream SSM method as a baseline. With NEZHA as the pre-trained model, baseline encodes address pairs by the model and uses Bi-LSTM to extract features to classify. In order to verify the importance of learning and leveraging hierarchical information in address matching, based on the baseline, we add a address resolution module to obtain hierarchical information, replace single word mask strategy with WWM strategy for address learning, and add another Bi-LSTM for element matching.

### 4.3 Evaluation method and hyper-parameter

Through F1, precision, accuracy and recall, we measure results of the address element resolution module. As for the other two modules, in order to evaluate the performance of our methods, we plan to use F1, accuracy and recall.

The hyper-parameters of the three modules are shown in Table 4.

**Table 4.** Hyperparameters of Three Module

| Hyperparameter | Address resolution module | Address learning module | Address matching module |
|---|---|---|---|
| batch | 128 | 32 | 32 |
| epoch | 100 | 150 | 3 |
| optim | sgd | Adam | AdaBelief |
| learning-rate | 6e-4 | 6e-5 | 5e-5 |
| seq-length | determined by the data | 100 | 200 for splicing whole address, 40 for splicing elements |

Note: This is hyper-parameter in our experiment.

### 4.4 Ablation experiment

The progressiveness of our method is assumed in two aspects: the learning of address hierarchical structure and fine-grained element, and the application of hierarchies in address matching. To verify the the assumption, we take ablation experiments, where we replace the whole word mask by single word mask strategy learning hierarchical information and cancel elements level matching using hierarchical information, respectively. On the basis of our method, we can cancel the two parts to verify their respective functions.

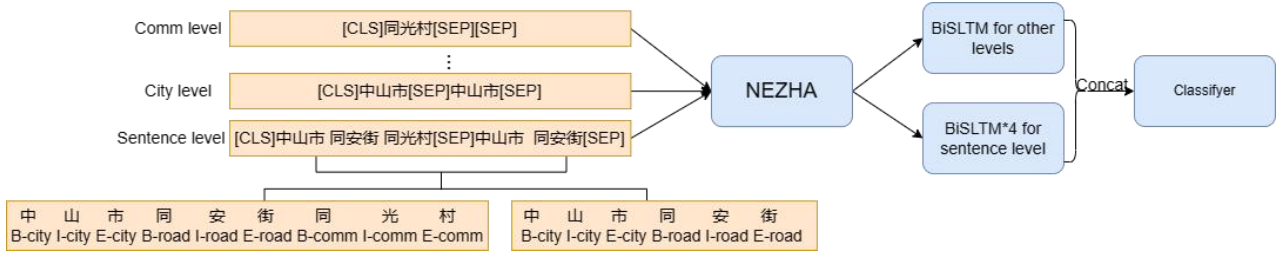

Fig.4. Structure of address matching module.

### 4.5 Result and analyse

Fig.5 shows result of address resolution in our method. From Fig.5, we can perceive that after about the 40th epoch, the model tends to converge with a great result that f1, pre, acc, and rec are all higher than 0.9 which can provide reliable support for subsequent modules.

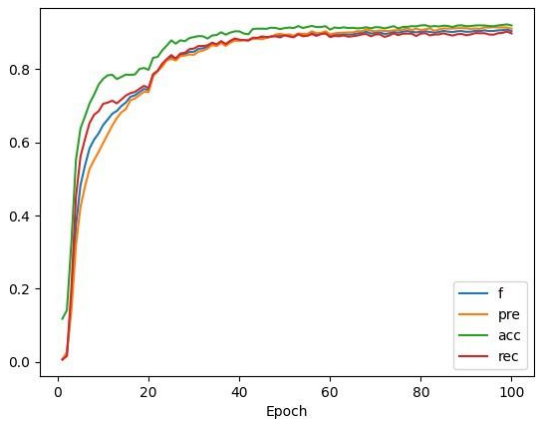

Fig.5. Result of address resolution.

Table 5 shows all comparison results of baseline, our proposed method and the two ablation experiments. Passing 3.23% F1 gap, improvement of accuracy of 1.77% and recall of 3.64% between baseline and our method, we detect that our method learning and using hierarchical information shows superiority over deep learning-based methods. There are two possible reasons: 1) The semantic representation of the address is more accurate, which helps to measure the similarity of the address and 2) After element matching, it can effectively deal with element missing, redundancy and other problems. The other results in Table 5 describe the impact of the above reasons.

Table 5. Results of Experiments

| Method | F1 | Acc | Recall |
| --- | --- | --- | --- |
| baseline | 82.10% | 86.49% | 81.01% |
| Our method | 85.33% | 88.26% | 84.65% |
| Our method without address learning | 82.55% | 85.63% | 81.78% |
| Our method without element matching | 84.45% | 87.95% | 84.36% |

Note: This is results of our experiment. Our method without address learning means using single word mask strategy instead of whole word mask strategy in address learning module. Our method without element matching means matching address only from the sentence level.

As shown in Table 5, Our method without address learning is a inferior to the state-of-the-art method with respect to F1, accuracy and recall, i.e. 2.78% in F1, 2.63% in accuracy and 2.87% in recall. It indicates that learning hierarchical structure and fine-grained element is more appropriate for element missing address and so on, for contributing association between elements in adjacent hierarchies. This is result of address learning learns the hierarchical sequence structure of address and the constraint relationship of fine-grained elements. Take the address" 江苏省南京市栖霞区栖霞街道栖霞街 60 号东巷口商住楼 2 栋 1 楼楼 (Floor 1, Building2, Dongxiangkou Commercial and Residential Build-ing, No. 60, Qixia Street, Qixia District, Nanjing City, Jiangsu Province)" and " 栖霞街 60 号 2 幢 106 室秦宁简餐 (Qinning Snack Restaurant, Room 106, Building 2, No. 60, Qixia Street)in the test set as an example, where the second address misses some address elements, it is predicted to be mismatch by using our method without address learning, while it is predicted correctly by using our method. We analyze the reason and find that learning hierarchical structure can memorize the relationship between address hierarchical structure and the sequence structure of address hierarchies.

As Table 5 depicts, Our method without element learning is a inferior to our method with respect to F1, accuracy and recall, i.e. 0.88% in F1, 0.31% in accuracy and 0.29% in recall. It demonstrates that due to comparison based on different hierarchies, our method is effective in address with hierarchical structure. Element matching brings better result since it is designed specifically for hierarchical structure. For some address, similar representation in different hierarchies will trigger matching at different hierarchies which cause error, since this improvement can relieve this problem.

" 仙岳路仙岳医院住院部 3 楼信息科 (Information Department on the 3rd floor of the inpatient department of Xianyue Hospital, Xianyue Road)" and " 仙岳路 1739 号厦门中医院内厦门中医院住院部 (Inpatient Department of Xiamen Traditional Chinese Medicine Hospital, No. 1739, Xianyue Road)" are predicted to be partial match via our method without element matching while are predicted to be similar to true label by our method. Comparisons such as " 仙岳路 (Xianyue Road)" with " 仙岳路 (Xianyue Road)" in road, " 仙岳医院 (Xianyue Hospital)" with " 厦门中医院 (Xiamen Traditional Chinese Medicine Hospital)" in poi are the keys to solving this pair of addresses which can be achieved by element matching. Furthermore, to relief the error caused by address element resolution with F1 in 90%, our method not only focus on element matching,but also sentence matching.

According to above analyse, it shows that address structure learning improves the address matching task more than element matching does in our method.

## 5 Conclusion

In order to better solve the address matching problem, we propose an address matching method by leveraging the address elements hierarchical information. The address composition obtained by the address element resolution model not only meets the fine-grained learning needs of the address, but also makes the comparison in the same hierarchy make up for the deficiency of the sentence-level comparison. As the experiments shown, this study strengthens the idea using hierarchical information in address matching and is first comprehensive fusion of two traditional address matching method. With the lack and injection of external address data ,the future work is a cross-national study about how to obtain and introduce external address knowledge for same address in completely different ways of writing.